\documentclass[conference]{IEEEtran}
\IEEEoverridecommandlockouts
\usepackage{cite}
\usepackage{booktabs}
\usepackage[colorlinks=true,urlcolor=blue,linkcolor=blue,filecolor=blue,citecolor=blue]{hyperref}
\usepackage{listings}
\usepackage{amsmath,amssymb,amsfonts,wasysym}
\usepackage{algpseudocode}
\usepackage{color,soul}
\usepackage{graphicx}
\usepackage{textcomp}
\usepackage{xcolor}
\def\BibTeX{{\rm B\kern-.05em{\sc i\kern-.025em b}\kern-.08em
    T\kern-.1667em\lower.7ex\hbox{E}\kern-.125emX}}

\DeclareMathOperator*{\argmin}{argmin}

\definecolor{codegreen}{rgb}{0,0.6,0}
\definecolor{codegray}{rgb}{0.5,0.5,0.5}
\definecolor{codepurple}{rgb}{0.58,0,0.82}
\definecolor{backcolour}{rgb}{0.95,0.95,0.92}

\lstdefinestyle{mystyle}{
    backgroundcolor=\color{backcolour},   
    commentstyle=\color{codegreen},
    keywordstyle=\color{magenta},
    numberstyle=\tiny\color{codegray},
    stringstyle=\color{codepurple},
    basicstyle=\ttfamily\footnotesize,
    breakatwhitespace=false,         
    breaklines=true,                 
    captionpos=b,                    
    keepspaces=true,                 
    showspaces=false,                
    showstringspaces=false,
    showtabs=false,                  
    tabsize=2
}

\newcommand\hll[1]{%
  \bgroup
  \hskip0pt\color{red!80!black}%
  #1%
  \egroup
}

\begin{document}

\title{Audiogram Digitization Tool for Audiological Reports
\thanks{This project would not have been possible without the financial support provided by the Workplace Safety Insurance Board of Ontario.}
}

\author{\IEEEauthorblockN{Fran\c{c}ois Charih}
\IEEEauthorblockA{\textit{Systems and Computer Engineering} \\
\textit{Carleton University}\\
Ottawa, ON, Canada \\
francoischarih@sce.carleton.ca}
\and
\IEEEauthorblockN{James R. Green}
\IEEEauthorblockA{\textit{Systems and Computer Engineering} \\
\textit{Carleton University}\\
Ottawa, ON, Canada \\
jrgreen@sce.carleton.ca}
}

\maketitle

\begin{abstract}
Multiple private and public insurers compensate workers whose hearing loss
can be directly attributed to excessive exposure to noise in the workplace.  The
claim assessment process is typically lengthy and requires significant effort
from human adjudicators who must interpret hand-recorded audiograms, often sent
via fax or equivalent. In this work, we present a solution developed in
partnership with the Workplace Safety Insurance Board of Ontario to streamline
the adjudication process. We present a flexible and open-source audiogram
digitization algorithm capable of automatically extracting the hearing
thresholds from a scanned or faxed audiology report as a proof-of-concept. The
algorithm extracts most thresholds within 5 dB accuracy, allowing to 
substantially lessen the time required to convert an audiogram into digital
format in a semi-supervised fashion, and is a first step towards the automation
of the adjudication process. The source code for the digitization
algorithm and a desktop-based implementation of our NIHL annotation portal is publicly available on GitHub
\href{https://github.com/GreenCUBIC/AudiogramDigitization}{https://github.com/GreenCUBIC/AudiogramDigitization}.
\end{abstract}


\section{Introduction}

Noise-induced hearing loss (NIHL) is a common consequence of long-term exposure
to noise in the workplace. In fact, a Canadian study~\cite{feder2017} recently
found, through a series of over 3,500 interviews, that 42\% of respondents were
exposed to hazardous levels of noise in the workplace. Moreover, 20\% of
respondents who reported being asked to wear hearing protective equipment by
their employer admitted to not following this rule consistently.  It is
therefore not surprising that numerous occupational NIHL-related claims are
received by public and private insurance companies yearly.

The Workplace Safety Insurance Board of Ontario (WSIB) reports receiving several
thousands of NIHL-related claims every year which can take several months to
process. The audiological reports received must be carefully interpreted by
adjudicators who apply a series of rules to determine the eligibility of the
claim. This is a time-consuming process that contributes to the lengthy
adjudication process.

The audiogram is a critical component of a NIHL-related claim. An audiogram
plots the hearing \textit{threshold} (minimal perceivable amplitude) in dB
across a range of frequencies. Different standard audiological symbols are used
to indicate whether a hearing threshold correspond to the left or right
ear, whether the threshold was obtained through air or bone conduction, and
whether masking was used or not to prevent the non-test hear from hearing tones
delivered to the contralateral ear. The shape of this audiometric curve is pivotal in
establishing the etiology of the hearing loss. For instance, individuals with
NIHL tend to have a notch in their air and bone conduction audiometric curves
(worse hearing) between 3,000 and 6,000 Hz~\cite{rabinowitz2000}.

While audiometers are now fully capable of generating digital versions of
audiograms, many hearing professionals still plot audiograms by hand in
audiological reports. While these reports differ slightly in their layout and
content, they typically consist of a single page with one or two audiogram plots
(combined or separated by ear), a brief summary of the findings, and potentially,
results from other tests (\textit{e.g.}, tympanogram, \textit{etc.}).

Because the reports are received by fax or as image files sent electronically,
eligibility rules can only very rarely be directly applied to reports submitted
to insurance companies or compensation boards. As such, an algorithm capable of
automatically extracting hearing thresholds from the audiograms on scanned or
faxed audiological reports to add to the client's electronic record could
significantly reduce the time required to process a claim.

In this work completed in partnership with the WSIB, we present an
audiogram digitization algorithm that combines traditional image processing and deep
neural networks trained with transfer learning to digitize audiograms. We
outline how the training data for this algorithm were collected, the training
procedure, and show how the algorithm can be useful in the adjudication of
NIHL-related claims.

\section{Related Work}
  Automated audiogram acquisition and interpretation has been an active area of research over the past two decades. Several automated audiometry tools were developed and marketed to counterbalance the predicted gap between the availability and need for hearing professionals~\cite{Margolis2008}. For instance, multiple tablet-based audiometers were designed to make hearing assessment possible outside a soundproof booth~\cite{Swanepoel2013,Margolis2011,Thompson2015}. These mobile audiometers are now deployed worldwide to afford access to basic hearing assessments to people in remote locations where hearing professionals are few or absent. 

  To generate actionable insight, the audiograms generated with these automated tools must be interpreted by a trained hearing professional or some other thoroughly validated automated method. Useful information that can be extracted from audiograms include the type of hearing loss (\textit{e.g.}, sensorineural, conductive or mixed), the severity of the hearing loss, and whether there is a need for immediate or long-term intervention(s), for example. A number of audiogram interpretation models have been developed to address the challenge of automatically extracting such information from audiograms. One of the first well-known methods to achieve this is the AMCLASS system~\cite{Margolis2007}. It is a rule-based system that was designed by hand to classify the hearing loss by severity, symmetry, configuration and type of hearing loss. Several groups leveraged large audiological databases and clustering-based approaches to generate a set of ``canonical'' audiograms against which new ones can be compared~\cite{Bisgaard2010,Lee2010}. More recently, we described a data-driven model trained on hundreds of expert-annotated audiograms to take an audiogram as an input, and generate a concise summary of the severity, configuration and symmetry across ears of the hearing loss~\cite{Charih2020}. Crowson \textit{et al.} trained and compared several convolutional neural network architectures for the task of identifying the type of hearing loss from \textit{undigitized} audiogram images, achieving quite impressive results ($>$95\% accuracy)~\cite{Crowson2020}. It is unclear whether their model generalizes well, however, given that all the audiograms were generated at the same instution. They argue that bypassing the digitization step simplifies the process of extracting insight from audiograms, but this comes at a cost in terms of flexibility. For example, it makes it more difficult to classify audiograms based on a set arbitrary rules defined by an institution. It is true that a model trained on audiogram images can implicitely learn the rules from a large number of audiograms annotated with the rules, but this will not outcompete the perfect accuracy of digitized audiograms upon which classification rules are applied. Furthermore, if digitization is accurate on a wide variety of audiology reports, then subsequent audiogram interpretation tasks (\textit{e.g.}, hearing loss type identification, determination of the best intervention, hearing aid tuning, \textit{etc.}) become agnostic to the actual way the audiogram was plotted (\textit{i.e.}, handwritten, grid dimensions, image quality, etc.), and therefore require fewer audiograms to train.
  
  These automated audiogram classification models typically, though not always~\cite{Crowson2020}, rely on digital audiometry data to analyze the subject's hearing, hence the need for audiogram digitization tools capable of accurately turning images of audiograms into a list of thresholds are often needed. The only other method that we are aware of that can achieve this is the \textit{Multi-stage Audiogram Interpretation Network (MAIN)} of Li \textit{et al.}, 2021~\cite{Li2021}. The authors trained several convolutional neural networks to extract audiograms, symbols and axis labels from audiogram images, to finally reconstruct the audiogram digitally\footnote{The approach of Li \textit{et al.}, 2021 shares striking similarities with ours, and was uploaded to the arXiv pre-print server in the time between the drafting of this paper and its submission. We were not aware of it as we were developing our own method.}. There are a couple limitations associated with MAIN. First, it can only process unmasked air conduction thresholds, \textit{i.e.}, only two of the eight commonly encountered types of measurements. Second, it was trained and tested on a small dataset called \textit{Open Audiogram} consisting of 64 unique audiograms, all generated with the same system. Multiple photos of each audiogram were taken in different conditions to augment it. MAIN will require validation on a larger and more diverse dataset to demonstrate its generality.

\section{Digitization Strategy}

\begin{figure}[]
  \centering
  \includegraphics[width=0.8\linewidth]{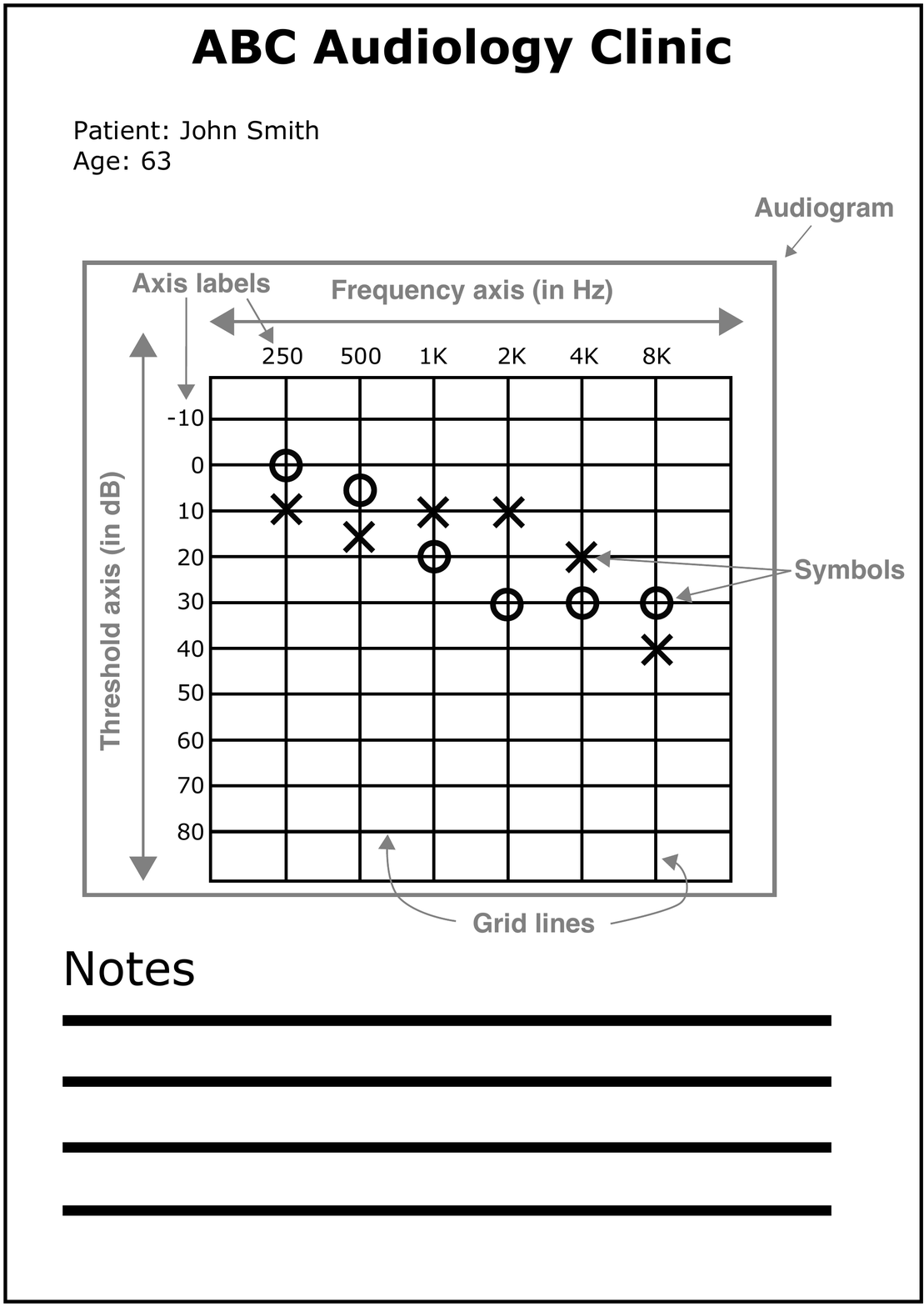}
  \caption{\textbf{Components of interest within a typical audiology report.} 
  A typical audiology report contains general information about the patient,
  the audiogram, and a section for notes. Occasionally, the report will also
  contain tympanometry data in tabular or graphical form. The elements of interest
  that are extracted for the purpose of this work are identified in gray.
  }%
  \label{fig:simple-report}
\end{figure}

In order to extract hearing thresholds from an audiological report, the following
elements (shown in \autoref{fig:simple-report}) must be located and identified:

\begin{enumerate}
  \item \textbf{Audiogram:} to identify the region(s) of interest in the report.
  \item \textbf{Axis labels and grid lines:} important for the conversion of pixel coordinates to frequency-threshold coordinates.
  \item \textbf{Audiological symbols:} the hearing thresholds of the individual.
\end{enumerate}

A few techniques were considered to tackle the object detection problem.
For instance, we initially
considered applying template matching to locate the audiological symbols within
the report, but rapidly realized that several problems made this approach
less than ideal. First, the symbols are often drawn by hand, and a given symbol
(\textit{e.g.}, a cross) will be drawn differently depending on the person recording the
audiogram. Next, overlapping symbols are a frequent occurrence, and as such, it
is not uncommon to see 2 or 3 overlapping symbols. For these reasons, the number
of templates that would be required to handle all possible symbols and symbol
combinations would be prohibitively large.

\begin{figure}[]
  \centering
  \includegraphics[width=1\linewidth]{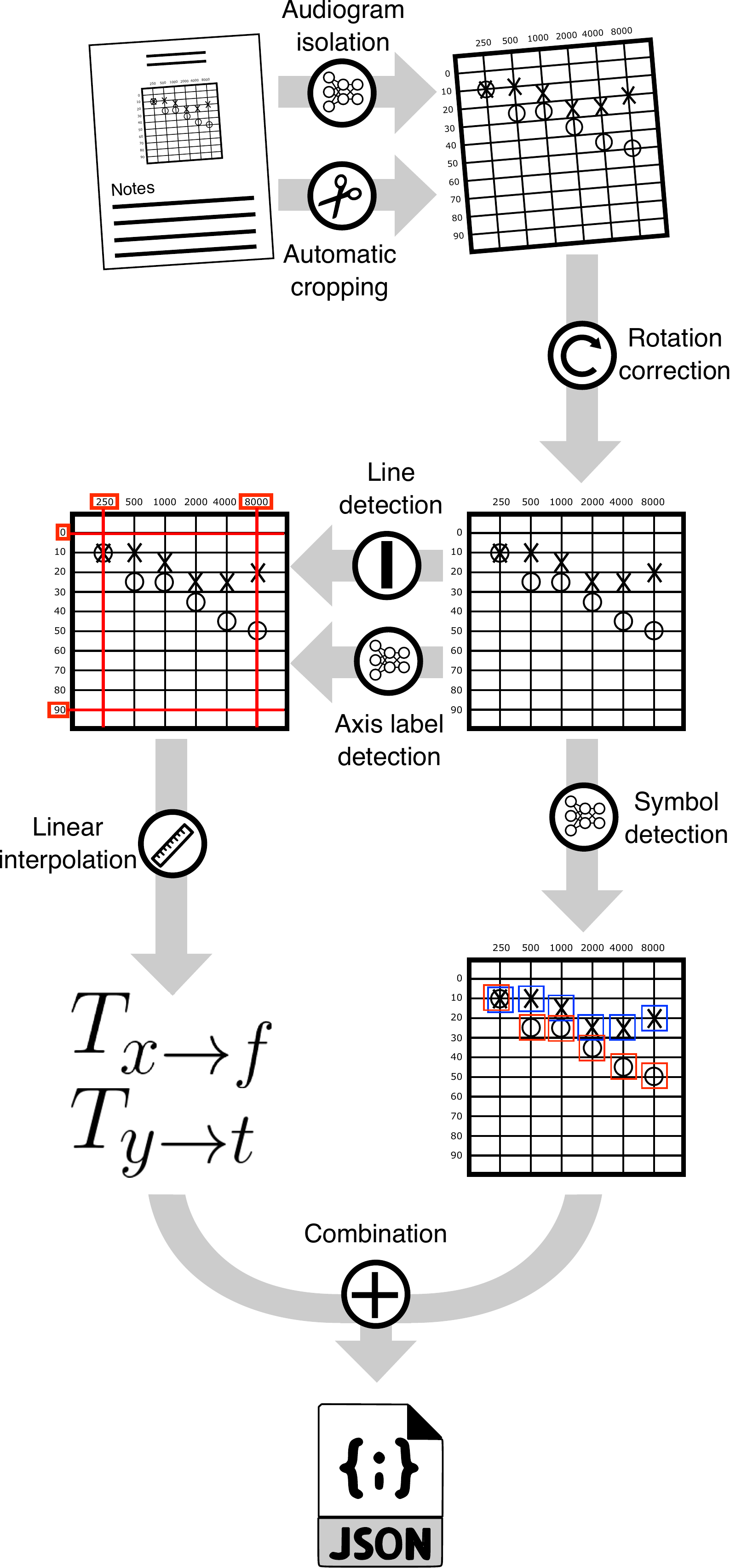}
  \caption{\textbf{Audiogram digitization strategy.} Our strategy combines a series of deep learning-based object detection tasks with traditional image processing procedures (line detection, cropping, rotation). These steps enable the derivation of pixel-to-frequency and pixel-to-threshold transforms through linear interpolation. These transforms are subsequently applied to the pixel coordinates of detected audiological symbols to generate a JSON document listing the individual hearing thresholds.}%
  \label{fig:strategy}
\end{figure}

We finally settled for the strategy illustrated in \autoref{fig:strategy}.
This strategy combines transfer learning (object detection) and traditional
image processing techniques. Pre-trained deep convolutional neural networks
are fine-tuned for detection of audiograms, axis labels, and symbols within a
report. Line detection techniques are applied to correct for audiogram rotation
and to compute the pixel-to-frequency/threshold transform. The algorithm
is described in greater details later in this paper.

\section{Data Collection}

Knowing that transfer learning would be employed to train object detectors for
elements composing the audiological report, we sought to assemble a dataset of
several thousands of annotated audiological reports.

The WSIB provided a large dataset of approximately 3,200 anonymized audiological
reports from claims received in 2006 or later. The reports all consisted of a
single page on which the audiograms were drawn in one or two plots. A
qualitative assessment of the dataset revealed a highly heterogeneous dataset
where reports varied greatly in terms of layout, resolution, handwriting, and
completeness.

To collect the annotations required to develop and evaluate the strategy
presented previously, we developed the \textit{NIHL Portal}, a web portal
specifically tailored to collect the required data rapidly and ergonomically.
The user interface was developed using React.js (\href{https://reactjs.org}{https://reactjs.org}) while its
backend was implemented using Flask (\href{https://flask.palletsprojects.com}{https://flask.palletsprojects.com}) as the
server and PostgreSQL (\href{https://www.postgresql.org/}{https://www.postgresql.org/}) as the database.

\begin{figure}
\begin{lstlisting}
  [
    {
      "boundingBox": {
        "x": number, 
        "y": number, 
        "width": number,
        "height": number
      },
      "corners": [
        {
          "x": number,
          "y": number,
          "position": { 
            "vertical": string,
            "horizontal": string
          },
          "frequency": number,
          "threshold": number,
        }, ...
      ],
      "labels": [
        {
          "boundingBox": {
            "x": number,
            "y": number,
            "width": number,
            "height": number
          },
          "value": string
        }, ...
      ],
      "symbols": [
        {
          "boundingBox": {
            "x": number, 
            "y": number,
            "width": number,
            "height": number
          },
          "response": boolean, 
          "measurementType": string 
        }, ...
      ]
    }, ...
  ]
  \end{lstlisting}
\caption{\textbf{Schema of a JSON annotation produced by the NIHL Portal.}
Annotations are stored as an array of one or two items, one for each
audiogram in the report. The locations of the audiogram bounding box,
corners of the audiogram, axis labels, and symbols are collected.}
\label{fig:annotation-schema}
\end{figure}

\begin{figure*}[ht]
  \centering
  \includegraphics[width=0.9\linewidth]{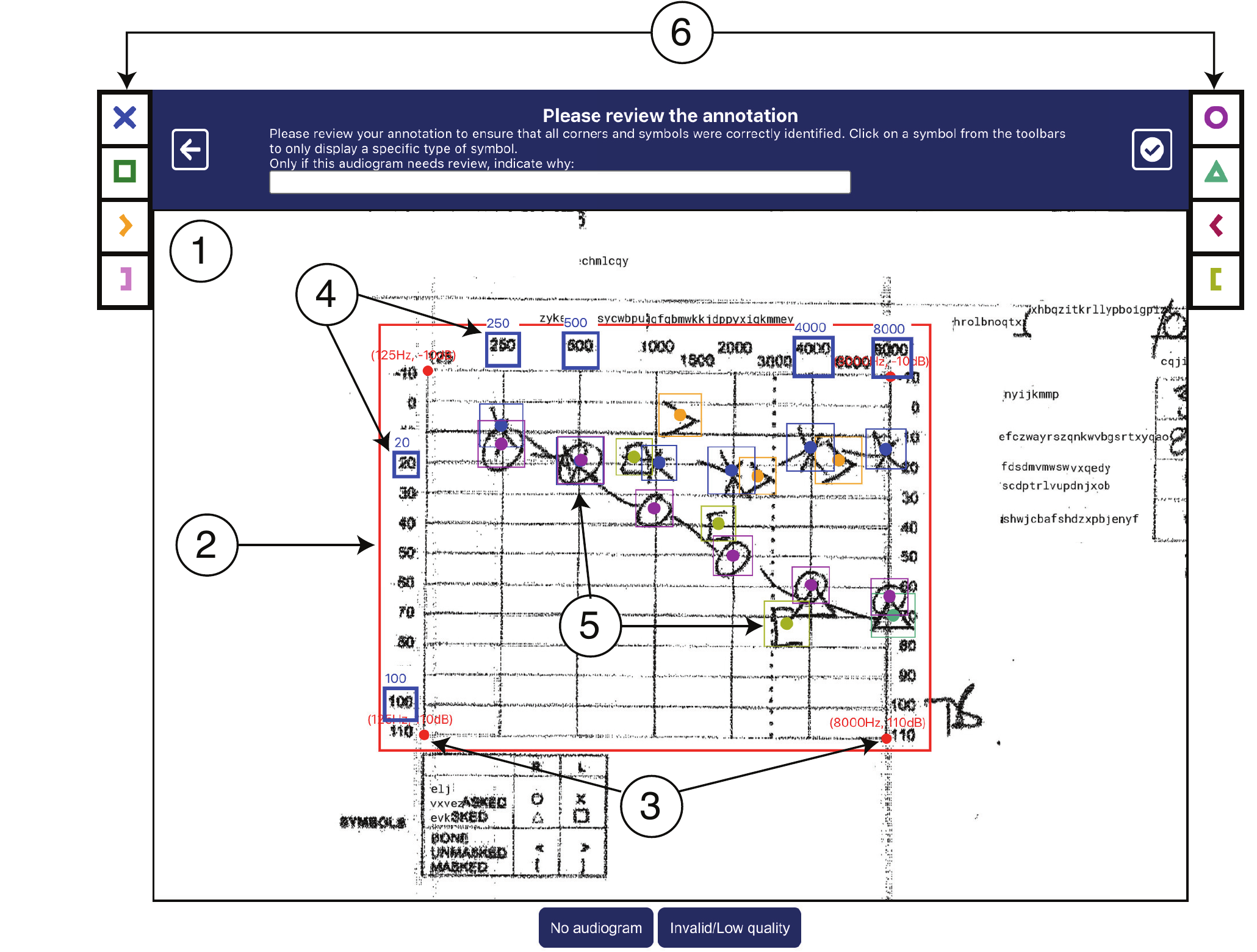}
  \caption{\textbf{NIHL Portal.} The annotator is shown the image of an audiological report in the viewport (1) and is asked to draw the bounding box around the audiogram (2). The annotator then proceeds to annotate the outermost corners (3), the axis labels (4), and the audiological symbols (5) by selecting the appropriate symbols from a tool bar (6).}%
  \label{fig:figures/portal}
\end{figure*}

Annotators were asked to draw, using the interface shown in
\autoref{fig:figures/portal}, bounding boxes for the audiogram, the axis labels,
and the individual threshold symbols. Annotators were also asked to indicate the
location of the audiogram corners along with the corresponding
frequency/threshold coordinates, so that a ground truth audiogram could be
computed from the annotation. The annotation process was completed over a period
of approximately 4 months. The JSON schema of the annotations collected is presented
in \autoref{fig:annotation-schema}.

\section{Digitization Algorithm}

\subsection{Audiogram Detection Model and Rotation Correction}

We used the YOLOv5s architecture developed by Ultralytics~\cite{ultralytics2020}
pre-trained on the COCO dataset~\cite{lin2015} to train an audiogram detection
model. The main purpose of this component of the digitization algorithm is to
isolate the audiogram(s) in the report so as to: 1) determine whether there
are audiograms in the report and 2) prevent the detection of symbols that are
outside the bounds of an audiogram plot.

The audiogram detection model was fine-tuned as per the instructions described
by the architecture's authors~\cite{ultralytics2020} on a computer with a NVIDIA
P100 GPU and 64G RAM. We used 80\% of the 3,000 reports for training, while the
remaining 20\% was used for validation. We set 206 reports aside for testing.
 The model was trained for 100 epochs, and
we selected the model that had the lowest generalized intersection over union
(GIoU)~\cite{rezatofighi2019} in validation. The procedure was repeated 3 times
with different seeds to emulate 3-fold cross-validation, allowing us to better estimate
the performance of the model and to quantify the uncertainty in that estimate.

When deployed, this model is used to crop the image of a report around the
audiogram(s). It is often the case that the orientation of the audiogram plot
must be corrected to make its grid lines horizontal and vertical. To this
end, we framed the problem of rotation correction as an optimization one. 

The first step of the procedure involves detecting lines in the audiogram with
the conventional Hough Transform~\cite{duda1972}. Lines that do not intersect
another line roughly perpendicularly ($\pm 1^\circ$) are not considered by the
algorithm, as they are unlikely to belong to the audiogram grid.  The correction
angle, $\theta_{corr}$, is then obtained by minimizing the sum of the deviations
of all lines from the horizontal (0$^\circ$) or vertical axis (90$^\circ$),
whichever is closest in terms of angle:

\begin{multline}
  \displaystyle{\argmin_{\theta_{corr}}} \left( \sum_{v \in V} |90^\circ - (\theta_v + \theta_{corr})|
  + \sum_{h \in H} |(\theta_h + \theta_{corr})| \right)
\end{multline}

where $V$ is the set of lines assumed to be vertical in the actual, unaltered report, i.e. $\theta_v \in (45^\circ, 135^\circ)$, and $H$ is the set of all other lines, assumed to be horizontal in the original, unaltered report, i.e. $\theta_h \in (-45^\circ, 45^\circ)$.

\subsection{Audiological Symbol Detection}

The audiological symbol detection model was trained similarly. We used 2803 \textit{audiograms} (not reports) for training/validation, and 273 audiograms for testing, and repeated the experiment 3 times with different seeds. We defined a total of 8 different classes corresponding to different audiological symbols accounting for the 4 different types of measurement (\autoref{table:symbols}).

\begin{table}[h]
\centering
\caption{Relevant audiological symbols considered in this study}
\label{table:symbols}
\begin{tabular}{@{}lccccc@{}}
\toprule
  & \multicolumn{2}{c}{Measurement type} & & \multicolumn{2}{c}{Ear} \\  \cline{2-3} \cline{5-6}
  & Conduction         & Masking      &   & Left       & Right      \\ \cline{2-6}
1 &      Air           & No           &   &    $\times$        &     $\ocircle$       \\
2 &      Air           & Yes           &   &    $\square$        &     $\triangle$       \\
3 &      Bone          & No           &   &    \texttt{>}        &     \texttt{<}       \\
4 &      Bone           & Yes           &   &    \texttt{]}        &     \texttt{[}       \\ \bottomrule
\end{tabular}
\end{table}

Once trained, the model can detect the symbols and provide the pixel coordinates of the center of any detected symbol's bounding box. However,
to be clinically meaningful, these pixel coordinates must be converted into frequency-threshold pairs.

\subsection{Axis Label Detection}

Given that the audiogram grids vary slightly in layout from one report to another, we sought
to identify the axis labels so that they could be associated to the lines
that make up the audiogram grid. This association allows for the derivation
of pixel-to-frequency and pixel-to-threshold mappings. The derivation of
these transforms using linear interpolation requires a minimum of two
correct grid-line associations per axis. Ideally, these
grid-line associations are at the opposite ends of the axes.

We initially attempted to apply optical character recognition to detect axis
labels using Google's open-source Tesseract engine~\cite{google2020}, but failed
to obtain reliable results despite our best efforts to preprocess the images by
adjusting parameters such as the contrast and brightness or by applying
techniques like thresholding and dilation/erosion.  However, given that the set
of labels encountered is finite, we framed this as an object detection problem
to be solved in the same way as audiogram detection, and collected axis label
annotations for 506 \textit{audiograms} (not reports), and used 465 for training/validation and 41 for testing.
In the same fashion as for the other two predictors, the process was repeated
three times with different seeds.

We fine-tuned
the same YOLOv5s model on a dataset of audiograms with annotations for the following frequency axis labels:
250 Hz, 500 Hz, 4,000 Hz and 8,000 Hz. We also included classes for equivalent representations
of the same frequencies (\textit{e.g.}, ``0.25'', ``0.5'', ``4K'', ``8'', \textit{etc.}). The classes
for the decibel axis labels included ``20'', ``60'', ``80'' and ``100''. We replicated the procedure
described in the previous section to train the axis label detection model.

\subsection{Deriving and Applying the Pixel Domain-to-Audiological Domain Transforms}

While it is trivial for human interpreters to identify the coordinates of
thresholds in the \textit{audiological domain} (in terms of frequencies and
threshold values) by visual inspection of the symbol and the surrounding axes, this task is far from trivial for
computers, which operate in the \textit{pixel domain}. One could presume that
given enough annotated audiograms, deep neural networks could be trained to
extract this information from images of audiograms. However, given the wide
variety of audiograms one may encounter, all unique in terms of their layout,
font, size, aspect ratio, it is highly unlikely that a couple of thousands of
audiograms would suffice.

Fortunately, one can make use of visual landmarks in the audiogram to derive the
transforms that convert pixel values along the $x$ axis to frequencies (in Hz)
and pixel values along the $y$ axis as thresholds (in dB). The most relevant
landmarks are the axis labels and the audiogram grid lines. \textit{Isn't that
how we,  humans, interpret this type of visual information after all?}

To derive the pixel-to-threshold transform (i.e. the $y$ axis), we pair detected axis labels with
the horizontal line ($\pm 1^\circ$) that intersects it and that is closest to
the center of its bounding box. The lines are detected with the Hough
transform~\cite{duda1972}. From this, we may generate a sorted list of tuples of
the form

\begin{equation}
       \{(y_1, t_1), ..., (y_n, t_n)\}
\end{equation}

where $y_i$ is the $y$ coordinate (in pixels) of the $i$-th horizontal line and
$t_i$ is the threshold value (in dB) of the associated label.

Provided that we are able to make two such associations, a transform can be
derived. If more than two such associations exist, the ones that are farthest
apart are used for increased resolution. The transform $T_{y \rightarrow t}$ is
simply:

\begin{equation}
  T_{y \rightarrow t}(y) = t_1 + \frac{(t_n - t_1)(y - y_1)}{y_n - y_1}
\end{equation}

We may proceed similarly for the pixel-to-frequency transform along the $x$
axis, except that one must account for the logarithmic nature of the frequency
scale. To do so, we use the linear \textit{octave scale}, where 125 Hz is the
$0^{th}$ octave, 250 Hz is the $1^{st}$ octave, 500 Hz is the $3^{rd}$ octave,
and so on. We can convert the frequency to its octave value with the equation:

\begin{equation}
  o(f) = \frac{\ln(f/125)}{\ln(2)}
  \label{eq:freq-to-oct}
\end{equation}

where $f$ is the frequency and $o(f)$ is the octave value of frequency $f$.

Then proceeding similarly as for the pixel-to-threshold transform
derivation, we may generate a sorted list of tuples by associating $x$ axis
labels with the vertical lines ($\pm 1^\circ$):

\begin{equation}
  \{(x_1, o_1), ..., (x_n, o_n)\}
\end{equation}

where $x_i$ is the $x$ coordinate (in pixels) of the $i$-th horizontal line and
$t_i$ is the octave value of the associated frequency label.

Then, we may use linear interpolation to derive the pixel-to-frequency tranform
$T_{x \rightarrow f}$, which converts back the octave value generated into a
frequency value by applying the reciprocal of \autoref{eq:freq-to-oct}:

\begin{equation}
  T_{x \rightarrow f}(x) = 125\times 2^{o_1 + \frac{(o_n - o_1)(x - x_1)}{x_n - x_1}}
\end{equation}

The frequency-threshold pairs generated by applying these transforms to the
coordinates of detected symbols virtually always yield values that approximate
the clinical measurements. Knowing that thresholds are measured at a set of
standard octave or semi-octave frequencies and that thresholds are measured in
increment of 5, we may correct the measurement by ``snapping'' the frequency
value to the nearest standard octave or semi-octave and the threshold value to
the nearest increment of 5.  For instance, an initial measurement of 53.3 dB
at 1,136 Hz would be corrected to 55 dB at 1,000 Hz. The rounded measurements
can then be used to populate a list of thresholds in JSON format (\autoref{fig:digitized-schema}).

\begin{figure}
\begin{lstlisting}
  [
    {
      "frequency": number, // in Hz
      "threshold": number, // in dB
      "masking": bool,
      "ear": string, // "left" or "right"
      "conduction": string // "air" or "bone"
    }, ...
  ]
  \end{lstlisting}
\caption{\textbf{JSON schema of a digitized audiogram.}
The JSON document produced by the algorithm is a list of JSON objects that
describe the threshold of hearing, which ear was tested, the measurement type
(air or bone) as well as whether masking was used or not for each frequency
tested.}
\label{fig:digitized-schema}
\end{figure}

\section{Results and Discussion}

\subsection{Performance of Individual Report Component Detectors}

To evaluate each independent object detector, we used a 3-fold cross-validation-like strategy. We computed the mean recall,
mean precision and mean average precision at 0.5 IoU (mAP@0.5) for each model.
The metrics are averaged over the different classes
for the axis label and symbol models (\autoref{table:individual-performance}).

\begin{table}[h]
\centering
\caption{Performance of the individual object detection models in 3-fold cross-validation}
\begin{tabular}{@{}lccc@{}}
\toprule
            & \multicolumn{3}{c}{Model}       \\ \cline{2-4}
            & Audiogram & Axis Label & Symbol \\\hline
Recall &     $1.00 \pm 0.00$     &    $0.70 \pm 0.12$        &   $0.83 \pm 0.04$     \\
Precision   &     $1.00 \pm 0.00$    &     $0.79 \pm 0.14$       &   $0.86 \pm 0.04$     \\
mAP@.5      &     $0.84 \pm 0.01$    &    $0.34 \pm 0.06$        &   $0.39 \pm 0.01$    \\ \bottomrule
\end{tabular}
\label{table:individual-performance}
\end{table}

Unsurprisingly, the easiest task is that of detecting audiograms within an audiology report.
The model trained to do so achieved a very high performance, with perfect
recall and precision.

Symbol detection was more difficult. Our
symbol detection model achieved an mAP@0.5 of $0.39 \pm 0.01$. The difficulty of this task
was not unforeseen, as the symbols are often hand drawn and overlapping, which leads to high variability
within a single report (by the same hearing clinician) and from one report to another
(between hearing clinicians).

It is the axis label detection model that achieves the lowest performance. While
it is true that the font used on the grids varies from one report to another and
that this may adversely affect performance, it seems more likely that the
scarcity of reports containing label annotations used in training (463) is responsible for this. 

\subsection{End-to-End Performance of the Digitization Algorithm}

To test the end-to-end performance of the digitization algorithm,
we ran the complete digitization algorithm on the 206 reports that were not used for training.
We computed the precision and recall of the digitization algorithm for every report
that could be successfully digitized as follows:

\begin{equation}
  Pr = \frac{|\mathcal{T}_{detected}\cap \mathcal{T}_{actual}|}{|\mathcal{T}_{detected}|}
\end{equation}

\begin{equation}
  Sn = \frac{|\mathcal{T}_{detected}\cap \mathcal{T}_{actual}|}{|\mathcal{T}_{actual}|}
\end{equation}

where $\mathcal{T}$ is a set of thresholds, each having a symbol (indicating the measurement type, see \autoref{table:symbols}),
a frequency, and a threshold value in decibels.

\begin{figure}[ht]
  \centering
  \includegraphics[width=1\linewidth]{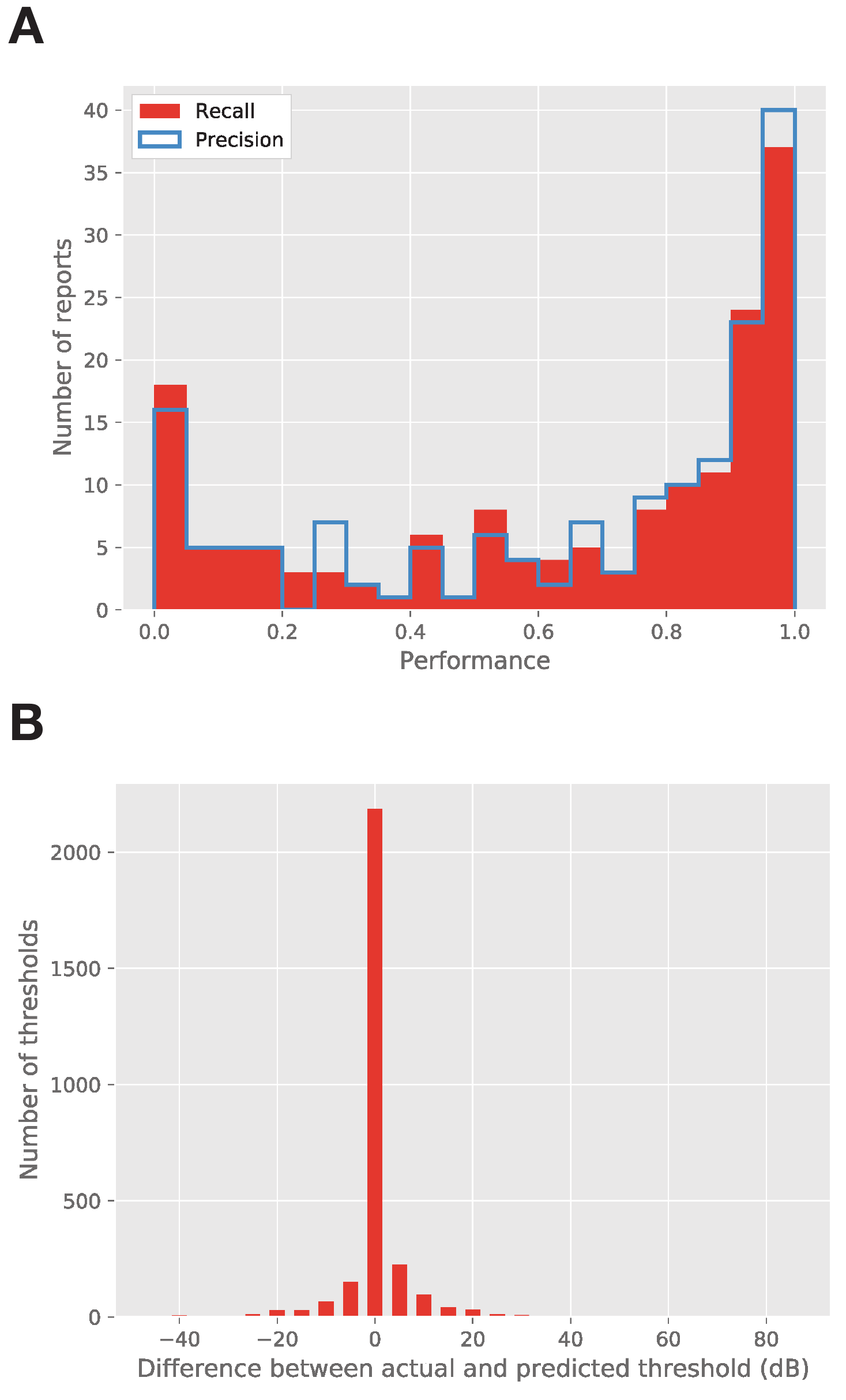}
  \caption{\textbf{Performance of the algorithm in end-to-end digitization.} (A)
  The recall and precision distribution of the algorithm over 163 successfully digitized reports. (B) The
  distribution of distances between the actual threshold values and those obtained via digitization.}%
  \label{fig:performance}
\end{figure}

The
digitization algorithm had a mean precision of $0.66 \pm 0.35$ and a mean
recall of $0.64 \pm 0.35$ over the remaining reports. The distribution
of precision and recall is shown in \autoref{fig:performance}A.  In
total, only 16 reports (7.8\%) were digitized perfectly. However, these metrics only
capture a small part of the story, as a threshold off by 5 dB is classified
incorrect while in many cases, it may not be clinically meaningful. One may
wonder why the recall and precision follow each other so closely. This is because the algorithm rarely fails to detect a symbol. It is far more common
for it to assign an incorrect threshold value to a symbol that actually exists. As
a result of the way the precision and recall are computed, if a symbol is given
the wrong threshold value, it will be counted both as a false positive
(a threshold that does not exist is predicted) \textit{and} a false negative
(actual threshold is missed).

Given that these metrics are strict and do not distinguish between digitized
values that are off from the actual value by 5 dB from those that are off by 30
dB, we computed the distribution of distances in dB between the actual and
digitized values (\autoref{fig:performance}B). The majority of the incorrectly extracted
thresholds (52\%) are off by only 5 dB.

Factors contributing to the difficulty of the task include the low resolution of
the images in our dataset. The fact that the digitization algorithm follows
largely a succeed-or-fail trend is mostly caused by a lack of robustness of the
grid extraction step which relies on the Hough transform and axis label
detection to determine the pixel-to-threshold and pixel-to-frequency transforms.
As mentioned previously, failure to properly derive the transforms leads to
inaccurate frequencies and/or threshold values for all the detected symbols
within a report. Moreover, hearing clinicians are inconsistent in how they
report bone conduction thresholds. The widely respected convention for air
conduction thresholds is to put the symbol directly on top of the vertical line
indicating a frequency. However, for bone conduction measurements some write
down the symbols for the left ear (\texttt{>} and \texttt{]}) to the right of
the frequency line and the symbols for the right ear (\texttt{<} and \texttt{[})
to the left of the frequency line, while others follow the same convention as
for air conduction thresholds. Errors in snapping these symbols to the correct
frequency adversely affects performance.

\subsection{Analysis of Digitization Failures}

\begin{figure*}[ht]
  \centering
  \includegraphics[width=1\linewidth]{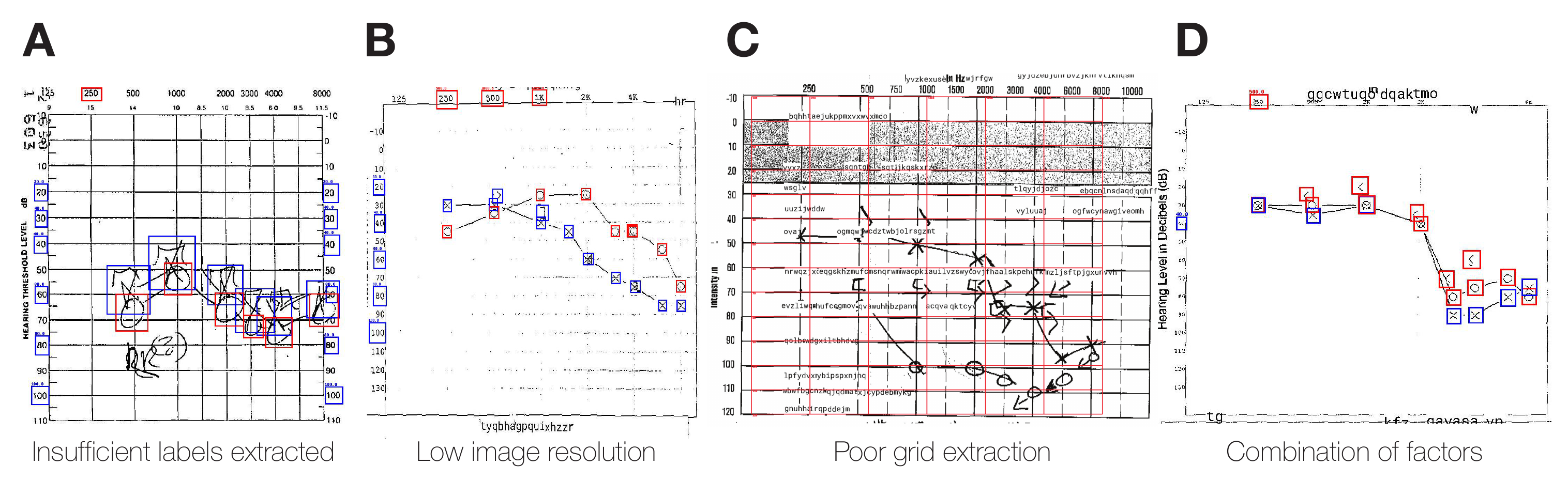}
  \caption{\textbf{Typical sources of audiogram digitization failure.} (A) An audiogram where only one axis label could be detected, leading to the inability to derive a pixel-to-frequency map. (B) An audiogram where the low scanned image resolution prevented the detection of grid lines. (C) Audiogram where the grid could not be accurately extracted (along the frequency axis), likely because of the fuzzy region denoting normal hearing between 0 and 20 dB. (D) An audiogram where multiple factors led to digitization failure, \textit{i.e.}, low resolution, absence of grid lines, and an insufficient number of axis label detections along the frequency axis.}%
  \label{fig:grid}
\end{figure*}

  Errors in one of the modules composing a multi-stage model may lead to
  inaccurate results or complete failure. The identification of bottlenecks or
  points of failure is key to understanding how to improve the overall model.
  
  The most critical step in the algorithm is arguably the derivation of
  accurate pixel-to-frequency and pixel-to-threshold transforms. These transforms
  rely on the accurate detection of at least 2 grid lines per axis and their
  associated labels.  This step is critical, because failure at this step will
  affect the conversion of \textit{all} symbol coordinates to
  frequency-threshold pairs downstream. Poor image resolution of the image and
  the variety of font types used to label the axes make this step the most
  challenging. In fact, of the 206 audiology reports used for testing, there were 43
  (21\%) for which the grid could not be extracted. This occurred when the
  transforms could not be derived because too few axis labels (\autoref{fig:grid}A), or too few grid lines were
  detected in images of poor resolution (\autoref{fig:grid}B). These two scenarios typically led to complete failure, \textit{i.e.}, prevented \textit{any} threshold from being extracted, even though the symbols are correctly detected. Occasionally, grid lines were incorrectly extracted (\autoref{fig:grid}C), leading to inaccurate digitization. Finally, the algorithm failed on reports where multiple of these sources of failure occurred; for example, when the resolution of the report scan was so low that neither the axis labels nor the grid lines could be extracted correctly (\autoref{fig:grid}D).

\subsection{Comparison with MAIN}

  We ran the pre-trained MAIN digitization model developed by Li \textit{et al.}~\cite{Li2021} on the same 206 audiology reports. It was unable to extract \textit{any} thresholds for 116 (56\%) of these reports. This
  occurred more than twice as frequently as for our own method (21\%). We also observed that, in contrast to our
  own method, theirs does not include logic to deal with duplicate detections, \textit{i.e.}, two or more
  thresholds detected for the same ear and frequency. For this reason, many of the reports that were in fact digitized contained up
  to three or four hearing threshold values detected for the same ear and frequency. This makes a fair assessment of precision
  and recall virtually impossible.

  These observations indicate that MAIN did not fully generalize to our dataset. This was reasonably foreseeable, given that it was trained on a relatively small dataset of audiograms generated using a single audiology software system. 

\begin{figure}[ht]
  \centering
  \includegraphics[width=1\linewidth]{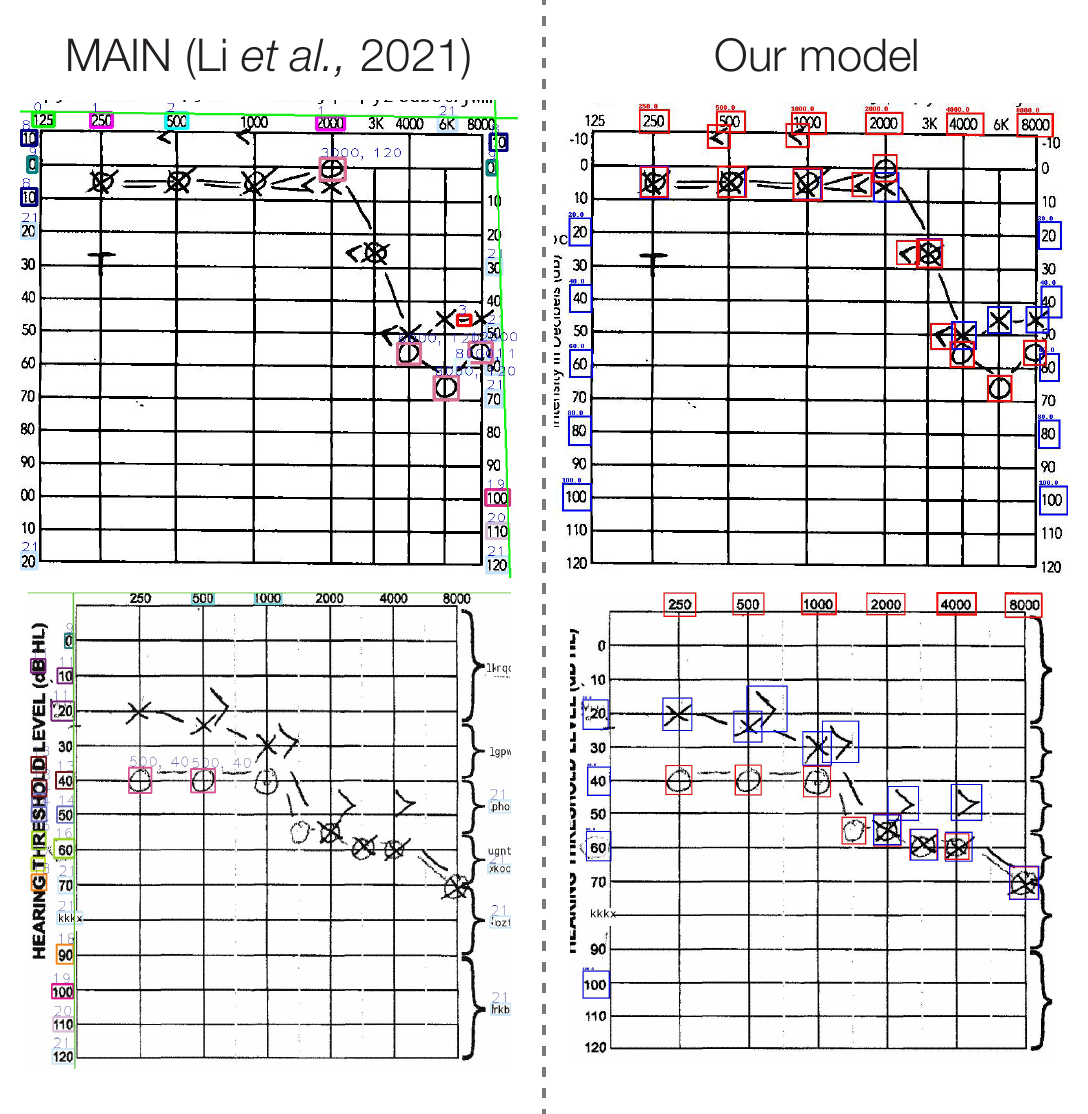}
  \caption{\textbf{Comparison of the components detected within representative audiograms.} The axis labels and audiological symbols detected by Li \textit{et al.}'s model (left) and ours (right) within two representative audiograms used for testing. The colors of the boxes on the left represent the different classes of objects.
  For our model, blue boxes represent label detections on the hearing loss axis or left ear symbols, while red boxes
  represent label detections on the frequency axis or right ear symbols.}%
  \label{fig:representative_comparison}
\end{figure}

  Upon closer inspection of the results, we observed that their symbol detection network failed to detect several symbols that it was trained to detect (\autoref{fig:representative_comparison}), \textit{i.e.}, the circle ($\ocircle$) and the cross ($\times$). The symbol detector appears to lack the ability to correctly detect symbols when they are overlapping. Their axis label detection network appeared to offer marginally better detection, but also failed to detect multiple axis labels. Clearly, the overall multi-stage model's performance is impaired by the low recall of its individual components.
  
 Given the similarity between the methods, it is likely that MAIN could perform significantly better if it were trained on a larger and more diverse set of audiology reports.

\section{Conclusion}

In this work, we presented a novel audiogram digitization algorithm capable of
extracting hearing thresholds from a variety of handwritten and
computer-generated audiology reports as a proof-of-concept. We have shown that
even if the task is seemingly simple, the variability in font, resolution,
handwriting, and conventions followed by hearing clinicians make the task quite
challenging.  We believe that the work that is most needed to improve the
performance of the algorithm involves refining the grid extraction step to
improve its robustness. Collecting additional annotations, especially for axis
labels, to retrain the axis label detector could also significantly improve the
performance of the algorithm. At this time, the algorithm lacks the accuracy to
be deployed in an unsupervised fashion, but combined with human supervision and
corrections, it can still drastically speed up the digitization process by
requiring only some manual adjustments to the extracted thresholds. The NIHL
portal bootstraps the audiogram digitization algorithm to produce an initial
annotation that can rapidly be adjusted by the annotator. This allows for the
expansion of the dataset to further train the audiogram, axis labels, and
symbol detection deep neural networks to enhance their accuracy.

Taken together, our work is novel as our model is the first of its kind to
be trained on a large dataset of faxed or scanned audiology reports and capable of
extracting hearing thresholds for all 8 commonly encountered audiological
measurements. In addition, we developed software that allows for the rapid
collection of annotations that can be used to further improve the recall and
precision of our individual component detectors.

We anticipate that this work will be of great interest not only to insurance
companies, who must process scanned or faxed audiological reports, but will also be
useful to researchers in the field of audiology interested in using
archived audiological records in paper format and to hospitals and clinics
in migrating records in their archives to a digital format. 

\section*{Acknowledgment}

The authors would like to thank Sherine Stephens, Valerie Hilton, Eric Gordon
and Ian Cross from the WSIB Innovation Labs for providing the data that was
used in this work. Furthermore, the authors also thank Jaser El-Habrouk, Abhinav Yalamanchili,
Ahmed Abdelrazik, Jason Fernandes, Tanzin Norjin, Syed Ahmed, Kenneth Calangan,
Brandon Ca and Vivian Han who each contributed numerous hours to annotate hundreds of audiological reports.

\bibliographystyle{IEEEtran}
\bibliography{wsib_paper}

% Generated by IEEEtran.bst, version: 1.14 (2015/08/26)
\begin{thebibliography}{10}
\providecommand{\url}[1]{#1}
\csname url@samestyle\endcsname
\providecommand{\newblock}{\relax}
\providecommand{\bibinfo}[2]{#2}
\providecommand{\BIBentrySTDinterwordspacing}{\spaceskip=0pt\relax}
\providecommand{\BIBentryALTinterwordstretchfactor}{4}
\providecommand{\BIBentryALTinterwordspacing}{\spaceskip=\fontdimen2\font plus
\BIBentryALTinterwordstretchfactor\fontdimen3\font minus
  \fontdimen4\font\relax}
\providecommand{\BIBforeignlanguage}[2]{{%
\expandafter\ifx\csname l@#1\endcsname\relax
\typeout{** WARNING: IEEEtran.bst: No hyphenation pattern has been}%
\typeout{** loaded for the language `#1'. Using the pattern for}%
\typeout{** the default language instead.}%
\else
\language=\csname l@#1\endcsname
\fi
#2}}
\providecommand{\BIBdecl}{\relax}
\BIBdecl

\bibitem{feder2017}
K.~Feder, D.~Michaud, J.~McNamee, E.~Fitzpatrick, H.~Davies, and T.~Leroux,
  ``Prevalence of {{Hazardous Occupational Noise Exposure}}, {{Hearing Loss}},
  and {{Hearing Protection Usage Among}} a {{Representative Sample}} of
  {{Working Canadians}}:,'' \emph{Journal of Occupational and Environmental
  Medicine}, vol.~59, no.~1, pp. 92--113, Jan. 2017.

\bibitem{rabinowitz2000}
P.~M. Rabinowitz, ``Noise-induced hearing loss,'' \emph{American Family
  Physician}, vol.~61, no.~9, pp. 2749--2756, May 2000.

\bibitem{Margolis2008}
R.~H. Margolis and D.~E. Morgan, ``Automated pure-tone audiometry: An analysis
  of capacity, need, and benefit,'' \emph{American Journal of Audiology},
  vol.~17, no.~2, pp. 109--113, Dec. 2008.

\bibitem{Swanepoel2013}
D.~W. Swanepoel, F.~{Maclennan-Smith}, and J.~W. Hall, ``Diagnostic pure-tone
  audiometry in schools: Mobile testing without a sound-treated environment,''
  \emph{Journal of the American Academy of Audiology}, vol.~24, no.~10, pp.
  992--1000, 2013 Nov-Dec.

\bibitem{Margolis2011}
R.~H. Margolis, R.~Frisina, and J.~P. Walton, ``{{AMTAS}}(\textregistered ):
  Automated method for testing auditory sensitivity: {{II}}. air conduction
  audiograms in children and adults,'' \emph{International Journal of
  Audiology}, vol.~50, no.~7, pp. 434--439, Jul. 2011.

\bibitem{Thompson2015}
G.~P. Thompson, D.~P. Sladen, B.~J.~H. Borst, and O.~L. Still, ``Accuracy of a
  {{Tablet Audiometer}} for {{Measuring Behavioral Hearing Thresholds}} in a
  {{Clinical Population}},'' \emph{Otolaryngology--Head and Neck Surgery:
  Official Journal of American Academy of Otolaryngology-Head and Neck
  Surgery}, vol. 153, no.~5, pp. 838--842, Nov. 2015.

\bibitem{Margolis2007}
R.~H. Margolis and G.~L. Saly, ``Toward a standard description of hearing
  loss,'' \emph{International Journal of Audiology}, vol.~46, no.~12, pp.
  746--758, Jan. 2007.

\bibitem{Bisgaard2010}
N.~Bisgaard, M.~S. M.~G. Vlaming, and M.~Dahlquist, ``Standard {{Audiograms}}
  for the {{IEC}} 60118-15 {{Measurement Procedure}},'' \emph{Trends in
  Amplification}, vol.~14, no.~2, pp. 113--120, Jun. 2010.

\bibitem{Lee2010}
C.-Y. Lee, J.-H. Hwang, S.-J. Hou, and T.-C. Liu, ``Using cluster analysis to
  classify audiogram shapes,'' \emph{International Journal of Audiology},
  vol.~49, no.~9, pp. 628--633, Sep. 2010.

\bibitem{Charih2020}
F.~Charih, M.~Bromwich, A.~E. Mark, R.~Lefran{\c c}ois, and J.~R. Green,
  ``Data-{{Driven Audiogram Classification}} for {{Mobile Audiometry}},''
  \emph{Scientific Reports}, vol.~10, no.~1, p. 3962, Dec. 2020.

\bibitem{Crowson2020}
M.~G. Crowson, J.~W. Lee, A.~Hamour, R.~Mahmood, A.~Babier, V.~Lin, D.~L.
  Tucci, and T.~C.~Y. Chan, ``{{AutoAudio}}: {{Deep Learning}} for {{Automatic
  Audiogram Interpretation}},'' \emph{Journal of Medical Systems}, vol.~44,
  no.~9, p. 163, Sep. 2020.

\bibitem{Li2021}
S.~Li, C.~Lu, L.~Li, J.~Duan, X.~Fu, and H.~Zhou, ``Interpreting {{Audiograms}}
  with {{Multi-stage Neural Networks}},'' Dec. 2021.

\bibitem{ultralytics2020}
\BIBentryALTinterwordspacing
G.~Jocher, A.~Chaurasia, A.~Stoken, J.~Borovec, NanoCode012, Y.~Kwon, TaoXie,
  K.~Michael, J.~Fang, imyhxy, Lorna, C.~Wong, Z.~Yifu, A.~V, D.~Montes,
  Z.~Wang, C.~Fati, J.~Nadar, Laughing, UnglvKitDe, tkianai, yxNONG,
  P.~Skalski, A.~Hogan, M.~Strobel, M.~Jain, L.~Mammana, and xylieong,
  ``{ultralytics/yolov5: v6.2 - YOLOv5 Classification Models, Apple M1,
  Reproducibility, ClearML and Deci.ai integrations},'' Aug. 2022. [Online].
  Available: \url{https://doi.org/10.5281/zenodo.7002879}
\BIBentrySTDinterwordspacing

\bibitem{lin2015}
T.-Y. Lin, M.~Maire, S.~Belongie, L.~Bourdev, R.~Girshick, J.~Hays, P.~Perona,
  D.~Ramanan, C.~L. Zitnick, and P.~Doll{\'a}r, ``Microsoft {{COCO}}: {{Common
  Objects}} in {{Context}},'' \emph{arXiv:1405.0312 [cs]}, Feb. 2015.

\bibitem{rezatofighi2019}
H.~Rezatofighi, N.~Tsoi, J.~Gwak, A.~Sadeghian, I.~Reid, and S.~Savarese,
  ``Generalized {{Intersection Over Union}}: {{A Metric}} and a {{Loss}} for
  {{Bounding Box Regression}},'' in \emph{2019 {{IEEE}}/{{CVF Conference}} on
  {{Computer Vision}} and {{Pattern Recognition}} ({{CVPR}})}.\hskip 1em plus
  0.5em minus 0.4em\relax {Long Beach, CA, USA}: {IEEE}, Jun. 2019, pp.
  658--666.

\bibitem{duda1972}
R.~O. Duda and P.~E. Hart, ``Use of the {{Hough}} transformation to detect
  lines and curves in pictures,'' \emph{Communications of the ACM}, vol.~15,
  no.~1, pp. 11--15, Jan. 1972.

\bibitem{google2020}
{Google}, ``Tesseract-ocr/tesseract,'' Nov. 2020.

\end{thebibliography}

\end{document}